\PassOptionsToPackage{numbers}{natbib}
\documentclass{article}

\usepackage[preprint]{neurips_2024}
\usepackage[utf8]{inputenc} % allow utf-8 input
\usepackage[T1]{fontenc}    % use 8-bit T1 fonts
\usepackage{hyperref}       % hyperlinks
\usepackage{url}            % simple URL typesetting
\usepackage{booktabs}       % professional-quality tables
\usepackage{amsfonts}       % blackboard math symbols
\usepackage{nicefrac}       % compact symbols for 1/2, etc.
\usepackage{microtype}      % microtypography
\usepackage{xcolor}         % colors
\usepackage{enumitem}

\usepackage{natbib}

\usepackage{amsmath}
\usepackage{graphicx}
\usepackage{authblk}

\makeatletter
\renewcommand{\@noticestring}{}
\makeatother

\title{AM‑Thinking‑v1: Advancing the Frontier of Reasoning at 32B Scale}

\begin{document}
\makeatletter
\renewcommand\@fnsymbol[1]{\ifcase#1\or 1\else\@arabic{#1}\fi}
\makeatother

\author{
  Yunjie Ji,\quad Xiaoyu Tian,\quad Sitong Zhao,\quad Haotian Wang,\\[0.3em]
  Shuaiting Chen,\quad Yiping Peng,\quad Han Zhao,\quad Xiangang Li
}

%\author{Yunjie Ji}
%\author{Xiaoyu Tian}
%\author{Sitong Zhao}
%\author{Haotian Wang}
%\author{Shuaiting Chen}
%\author{Yiping Peng}
%\author{Han Zhao}
%\author{Xiangang Li}

\affil{
    \raisebox{-0.4em}{\includegraphics[height=1.5em]{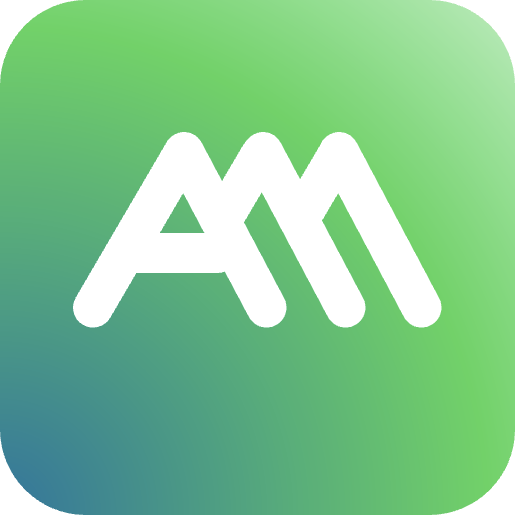}}
    \hspace{0.2em}a-m-team\thanks{The a-m-team is an internal team at Beike (Ke.com), dedicated to exploring AGI technology.}
}
\date{}

\maketitle

\begin{abstract}
%We present AM-Thinking-v1, a 32B dense language model that achieves state-of-the-art reasoning performance among open-source models of similar scale. Despite its compact size, AM-Thinking-v1 surpasses DeepSeek-R1 and rivals top-tier Mixture-of-Experts (MoE) models such as Qwen3‑235B‑A22B and Seed1.5-Thinking on reasoning-intensive benchmarks. Specifically, it achieves scores of 85.3 and 74.4 on AIME 2024 and AIME 2025, respectively, and 70.3 on LiveCodeBench, demonstrating robust capabilities in mathematical and coding tasks. Our results show that, starting from a fully open-source base model and reinforcement learning queries, a carefully designed post-training pipeline can elicit flagship-level reasoning from a 32B dense model. As model scales continue to grow, we argue that the 32B range remains a sweet spot for practical deployment and further fine-tuning. We hope our work inspires more research focused on unlocking the full potential of models at this scale—striking a balance between top-tier performance and real-world usability.

We present AM-Thinking-v1, a 32B dense language model that advances the frontier of reasoning, embodying the collaborative spirit of open-source innovation. Outperforming DeepSeek-R1 and rivaling leading Mixture-of-Experts (MoE) models like Qwen3-235B-A22B and Seed1.5-Thinking, AM-Thinking-v1 achieves impressive scores of 85.3 on AIME 2024, 74.4 on AIME 2025, and 70.3 on LiveCodeBench, showcasing state-of-the-art mathematical and coding capabilities among open-source models of similar scale.

Built entirely from the open-source Qwen2.5-32B base model and publicly available queries, AM-Thinking-v1 leverages a meticulously crafted post-training pipeline — combining supervised fine-tuning and reinforcement learning — to deliver exceptional reasoning capabilities. This work demonstrates that the open-source community can achieve high performance at the 32B scale, a practical sweet spot for deployment and fine-tuning. By striking a balance between top-tier performance and real-world usability, we hope AM-Thinking-v1 inspires further collaborative efforts to harness mid-scale models, pushing reasoning boundaries while keeping accessibility at the core of innovation.
% We have open-sourced our model on \href{https://huggingface.co/a-m-team/AM-Thinking-v1}{Hugging Face}.
We have open-sourced our model on \href{https://huggingface.co/a-m-team/AM-Thinking-v1}{Hugging Face}\footnote{\url{https://huggingface.co/a-m-team/AM-Thinking-v1}}.

\end{abstract}

\begin{figure}[h!]
    \centering
    \includegraphics[width=1.0\textwidth]{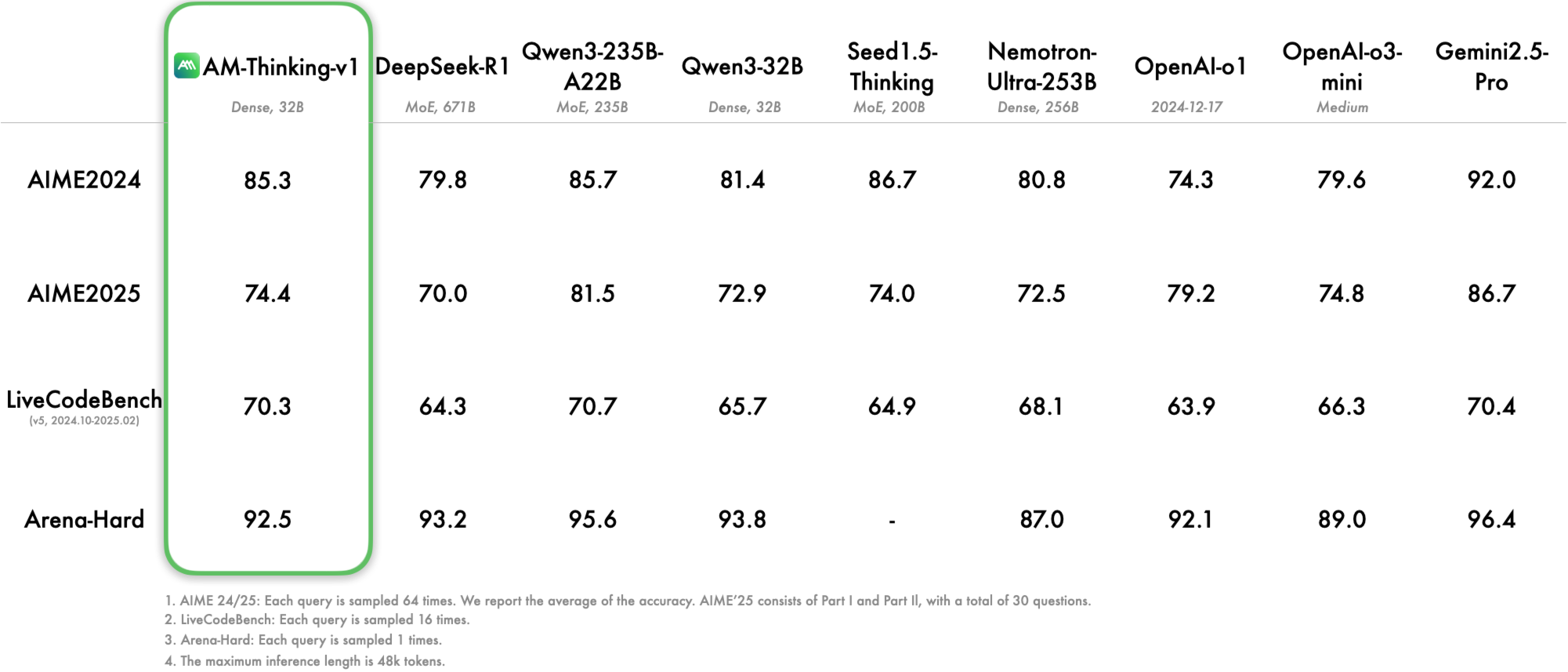}
    \caption{Comparison of Model Performance on Reasoning Benchmarks}
    \label{fig:benchmark}
\end{figure}

\section{Introduction}

Over the past six months, large language models (LLMs) have demonstrated remarkable improvements in reasoning, particularly in domains such as mathematical problem solving and code generation---tasks that require sophisticated logical inference. These advancements are expanding the practical applicability of LLMs across a broader range of real-world scenarios.

The release of DeepSeek-R1\citep{deepseekai2025deepseekr1incentivizingreasoningcapability} has shown that open-source communities are increasingly capable of building models that rival proprietary systems such as OpenAI’s o1\citep{openai2024reasoning}, Google’s Gemini 2.5\citep{googledeepmind2025reasoning}, and Anthropic’s Claude 3.7\citep{anthropic2025reasoning}. More recently, the emergence of Qwen3-235B-A22B\citep{qwen3} has further advanced the reasoning frontier of open-source models. However, many recent breakthroughs rely on extremely large-scale Mixture-of-Experts (MoE) architectures, which impose significant infrastructure burdens and make model deployment and fine-tuning considerably more complex.

In contrast, dense models of moderate size (e.g., 32B) offer better efficiency and deployability, yet often lag behind their MoE counterparts in reasoning performance. This contrast raises a critical research question: \textit{Can we unlock the reasoning potential of 32B-scale dense models---without relying on private data or massive MoE architectures---through a carefully designed post-training pipeline?}

To explore this question, we introduce and open-source AM-Thinking-v1, a reasoning-optimized language model built upon the publicly available Qwen2.5-32B\citep{qwen2, qwen2.5} base model. Our model achieves state-of-the-art performance among dense models of comparable size and even outperforms much larger MoE models in several reasoning benchmarks. Specifically, AM-Thinking-v1 achieves impressive scores of 85.3 and 74.4 on AIME2024\citep{maa_aime_2024} and AIME2025\citep{ye2025aimepreview}, two challenging math competition-style benchmarks, and 70.3 on LiveCodeBench\citep{jain2024livecodebench}, a widely used benchmark for evaluating code generation. 
It surpasses DeepSeek-R1 (671B MoE) and approaches or matches the performance of other top-tier MoE models such as Qwen3-235B-A22B and Seed1.5-Thinking\citep{seed2025seed15thinkingadvancingsuperbreasoning}, despite having only a fraction of their parameters.

Our success stems from a meticulously designed post-training framework that leverages publicly available training queries. We applied strict preprocessing to various open-source queries and instructions, including deduplication, removal of low-quality or multi-modal queries (e.g., those involving images), and thorough decontamination with respect to our evaluation benchmarks.
In particular, for mathematical queries—where we observed a high prevalence of noisy items—we constructed a comprehensive data processing pipeline that spans query filtering and ground-truth verification.

The post-training pipeline comprises two main stages: Supervised Fine-Tuning (SFT) and Reinforcement Learning (RL). Starting from Qwen2.5-32B base model, we apply SFT using a cold-start dataset that encourages a \textit{"think-then-answer"} pattern and builds initial reasoning capability. During RL, we incorporate difficulty-aware query selection and a two-stage training procedure to ensure both training stability and progressive improvement in performance.

In summary, AM-Thinking-v1 demonstrates that even without large-scale MoE architectures, dense models at the 32B scale can achieve reasoning capabilities comparable to the best available models. We hope this work serves as a practical reference for the community, highlighting how careful post-training design can bridge the performance gap while retaining the deployability advantages of moderate-scale models. This offers a promising direction for future research at the intersection of scalability, accessibility, and reasoning performance.

\section{Data}
All queries used in our training come from publicly available datasets. We begin by deduplicating the queries and filtering out low-quality ones. For mathematical queries with ground-truth answers, we further verify the correctness of the provided ground truth. Additionally, we filter model-generated responses based on quality and assign difficulty levels to each query based on the pass rate observed across multiple response attempts.

\subsection{Data Collection}
Our training data is collected from multiple publicly available open source datasets, spanning tasks such as mathematical reasoning, code generation, scientific reasoning, instruction follow, and general chat.

\textbf{Mathematical Reasoning }
During the collection of mathematical data, we ensure that each data point include a verifiable ground truth. We incorporate datasets such as OpenR1-Math-220k\citep{openr1}, Big-Math-RL-Verified\citep{albalak2025bigmathlargescalehighqualitymath}, data\_ablation\_full59K\citep{muennighoff2025s1simpletesttimescaling}, NuminaMath\citep{numina_math_datasets}, MetaMathQA\citep{yu2023metamath}, 2023\_amc\_data\citep{aops2023amc8p10}, DeepMath-103K\citep{deepmath}, and AIME\citep{di_zhang_2025}\footnote{AIME 2024 and AIME 2025 are excluded from the training data.}.

\textbf{Code Generation }
We ensure that all collected code data include verifiable test cases. Datasets selected for this category include PRIME\citep{yuan2024implicitprm}, DeepCoder\citep{deepcoder2025}, KodCode\citep{xu2025kodcode}, liveincode\_generation\citep{jain2024livecodebench}, codeforces\_cots\citep{penedo2025codeforces}, verifiable\_coding\citep{openr1_verifiable_coding_2025}, opencoder\citep{Huang2024OpenCoderTO}, OpenThoughts-114k-Code\_decontaminated\citep{openr1}, and AceCode-87K\citep{AceCoder}.

\textbf{Scientific Reasoning }
This category includes natural sciences (physics, chemistry, natural sciences) and logical reasoning. They primarily consist of multiple-choice questions, each paired with a reliable ground truth. We include datasets such as task\_mmmlu\citep{wang2022supernaturalinstructionsgeneralizationdeclarativeinstructions}, chemistryQA\citep{microsoft2021chemistryqa}, Llama-Nemotron-Post-Training-Dataset-v1\citep{nvidia2025llama3nemotron}, LOGIC-701\citep{hivaze_logic701_2023}, ncert\citep{NCERT_Physics_12th,NCERT_Physics_11th,NCERT_Chemistry_11th,NCERT_Chemistry_12th,NCERT_Biology_11th,NCERT_Biology_12th}, and logicLM\citep{longface2025logiclm}.

\textbf{Instruction Follow (IF) }
We select two instruction-following datasets: Llama-Nemotron-Post-Training-Dataset\citep{bercovich2025llamanemotronefficientreasoningmodels}, tulu-3-sft-mixture\citep{lambert2024tulu3}.

\textbf{General Chat}
This category includes a broad range of tasks, covering open-ended queries, general knowledge, and everyday reasoning, and it supports both single-turn and multi-turn interactions. The selected datasets are evol\citep{xu2024wizardlm}, InfinityInstruct\citep{InfinityInstruct2024}, open\_orca\citep{OpenOrca}, tulu-3-sft-mixture\citep{lambert2024tulu3}, natural\_reasoning\citep{yuan2025naturalreasoningreasoningwild28m}, flan\citep{longpre2023flan}, ultra\_chat\citep{ding2023enhancing}, and OpenHermes-2.5\citep{OpenHermes2.5}.

\subsection{Query filtering}
After collecting the data, we first remove duplicates, then apply two cleaning steps to address common query quality issues:

\begin{itemize}
    \item \textbf{Removal of queries containing URLs.}
Since the model cannot access external links during training, the presence of URLs may lead to hallucinations or misleading outputs.

    \item \textbf{Removal of image-referencing queries.}
Since our model is purely text-based, it cannot perceive or process any visual information; such queries are therefore excluded from training.
\end{itemize}
Finally, we remove queries from the training set that are similar to those in the evaluation set, using both exact matching and semantic deduplication.

\subsubsection{Mathematical query filtering}
During our analysis of the mathematical data, we identify issues with unclear or incomplete query descriptions and incorrect ground truths. To address the former, we use an LLM to analyze and filter out queries lacking clear or complete descriptions. For the latter, we implement a rigorous ground truth validation process: for each query, we prompt DeepSeek-R1\citep{deepseekai2025deepseekr1incentivizingreasoningcapability} to generate multiple responses  and compare the most frequent answer (Deepseek-R1-common) with the original ground truth using  \verb|math_verify|\footnote{https://github.com/huggingface/Math-Verify}. Discrepancies between model predictions and the original ground truth prompt us to re-evaluate the correctness of certain annotations. For these cases, we consult o4-mini\citep{openai2025reasoning} to obtain an alternative answer (o4-mini-answer). If \verb|math_verify| determines that o4-mini-answer and Deepseek-R1-common produce equivalent results, we consider the original ground truth potentially incorrect and revise it to o4-mini-answer.

Following this processing, we further identify and handle specific data types unsuitable for training: mathematical proof problems and queries with multiple sub-questions are filtered out. While multiple-choice questions are also deemed unsuitable, their significant volume prompt us to rewrite them as fill-in-the-blank questions instead of discarding them.

 % \textbf{Ground Truth }\textbf{Validation:} Our ground truth validation process for mathematical data involved distilling Deepseek-R1\citep{deepseekai2025deepseekr1incentivizingreasoningcapability} multiple times and extracting the content within the answer boxes for each run. The most frequent answer, termed Deepseek-R1-common, was then compared against the existing ground truth. If they passed our verification (\verb|math_verify|\footnote{https://github.com/huggingface/Math-Verify}), the ground truth was considered reliable. Otherwise, we flagged the ground truth as potentially incorrect. For these unreliable data points, we consulted o4-mini\citep{openai2025reasoning} to obtain an answer (o4-mini-answer). If o4-mini-answer and Deepseek-R1-common passed by \verb|math_verify|, we revised the original ground truth to o4-mini-answer. Through this rigorous approach, we adjusted the ground truth for tens of thousands of data entries.

\subsection{Synthetic response filtering}
After query filtering, we apply three methods to filter out low-quality synthetic response:

\begin{itemize}
\item  \textbf{Perplexity-based Filtering.} We use our previously trained 32B model\citep{zhao20251} to compute the perplexity (PPL) of each model-generated response. Responses with PPL scores exceeding a predefined threshold are discarded.
\item  \textbf{N-gram-based Filtering.} We discard model responses containing repeated phrases of a certain minimum length that appear consecutively.

\item  \textbf{Structure-based Filtering.} For multi-turn dialogues, we ensure that the final turn is an assistant response. Additionally, we require that each model-generated reply contains both a complete think and answer component.
\end{itemize}
For each query, multiple responses are generated. We then compute a verify score for every query to assess response quality or query difficulty. For queries with ground truth answers, we calculate the pass rate across the multiple generated responses. For queries without ground truth, we employ a large language model (LLM)-based reward model to score each response, and use the average score as the final verification signal.
The scoring procedure is detailed in Section~\ref{sec_reward}.

% Considering the varying learning needs across different training stages, we propose a dynamic data selection strategy. In the initial phases, we can train the model on a mix of both easy and hard data. However, in later stages, to further enhance model performance, it becomes crucial to identify and prioritize data with higher learning potential. To this end, we introduce the Coefficient of Variation (CV) as a metric. A higher CV value for a query indicates greater inconsistency in the model's performance across the distillations, suggesting that the model would benefit more from learning from such data.

\section{Reward}
\label{sec_reward}
\subsection{Verifiable Queries}
In the case of math, code, and instruction-following (IF) queries with available ground truth or test cases, we employ rule-based verification or code execution to assess the correctness of model responses.

\subsubsection{Math}
For mathematical queries, the reward is determined by verifying the model's final answer. The process begins by extracting the answer from the last boxed (\texttt{\{\}}) content of the model's answer content. This extracted answer is then validated against the reference using \verb|math_verify|\footnote{https://github.com/huggingface/Math-Verify}. This tool normalizes answers to handle different representations for comparison, outputting \texttt{true} on match or \texttt{false} otherwise. Reward score is 1 for a correct answer and 0 for an incorrect answer.
\subsubsection{Code}

For code queries equipped with predefined test cases, the verification process is executed within a secure code sandbox environment. This sandbox currently supports evaluation for multiple programming languages, including Python and C++.

\textbf{Code Segmentation Extraction } Extraction of code segmentation is facilitated by specific delimiters. For example, Python code blocks are identified by enclosing them within \texttt{\textasciigrave\textasciigrave\textasciigrave}\texttt{python} and \texttt{\textasciigrave\textasciigrave\textasciigrave}, while C++ code blocks utilize \texttt{\textasciigrave\textasciigrave\textasciigrave}\texttt{cpp} and \texttt{\textasciigrave\textasciigrave\textasciigrave}. This delimitation approach is similar to that commonly used in Markdown.

\textbf{Test Case}  The execution of code in our dataset primarily takes two forms: method call and standard input/output. For these two forms, two distinct test case types are employed. The details of each type are provided below:
\begin{itemize}
    \item \textbf{Method Call Test Cases:} These queries require the implementation of a specific method or function. Test cases for this type are defined by a particular function name, along with input values and their corresponding expected outputs. To facilitate efficient execution within the sandbox environment, these test cases are automatically converted into assertion statements.

    \item \textbf{Standard Input/Output Test Cases:} These queries that typically do not specify a distinct entry function are common in scenarios such as competitive programming or scripting tasks, where code reads from standard input and writes to standard output.For these queries, test cases are handled via standard input (\texttt{stdin}) and standard output (\texttt{stdout}). 
\end{itemize}

% The figure below illustrates examples of the two test case types:
\begin{figure}[ht]
\centering
\includegraphics[width=0.65\linewidth]{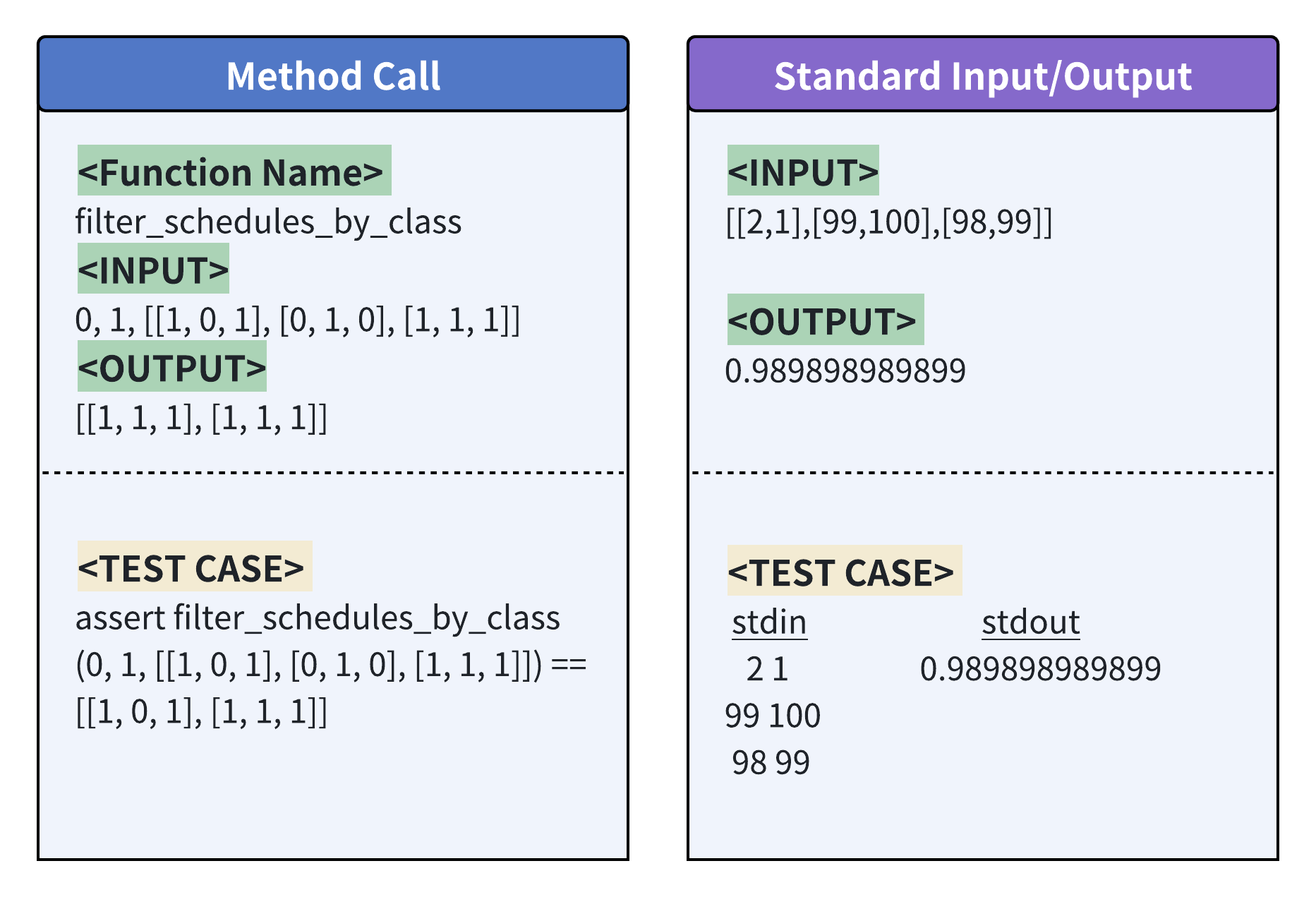}
\caption{Method Call And Standard Input/Output test case examples}
\label{fig:testcase}
\end{figure}

\textbf{Robust Cloud Sandbox Environment}  Meeting the dual requirements of secure code execution (mitigating malicious content risks) and robust performance under high concurrency (preventing unexpected timeouts) is central to our code sandbox design. This is achieved by deploying the sandbox as a distributed cloud service leveraging multiple machines. The distributed architecture, combined with load balancing and queue management, ensures both the isolation necessary for security and the capacity needed for reliable high-volume execution. 

For each code query with its typically associated multiple test cases, the final reward score is 1 if all tests pass, and 0 otherwise.
\subsubsection{Instruction Follow}
To obtain a reward score for instruction follow, we employ the \texttt{IFEval}\citep{zhou2023instructionfollowingevaluationlargelanguage} validator as our verifier. The validator receives instruction identifiers (\texttt{instruction\_id\_list}) and their arguments (\texttt{kwargs}). Here is an example:
% \begin{minted}[fontsize=\small]{json}
% {
%     "query": "What are some tips for managing stress during a pandemic?
%     Your response should contain at least 3 bullet points. 
%     Use the markdown bullet points such as: * This is point 1.
%     Also, your response should contain less than 100 words.
%     In 2008, a groundbreaking 9‐page PDF changed the money game: 
%     Bitcoin’s whitepaper. Authored by the mysterious Satoshi Nakamoto, 
%     it introduced a peer‐to‐peer electronic cash system bypassing central banks.",
%     "instruction_id_list": [
%                               "detectable_format:number_bullet_lists",
%                               "length_constraints:number_words"
%                            ],
%     "kwargs": [
%                 {"num_bullets": 3}, 
%                 {"num_words": 100, "relation": "less than"}
%               ]
% }
% \end{minted}
\begin{figure}[ht]
\centering
\includegraphics[width=0.95\linewidth]{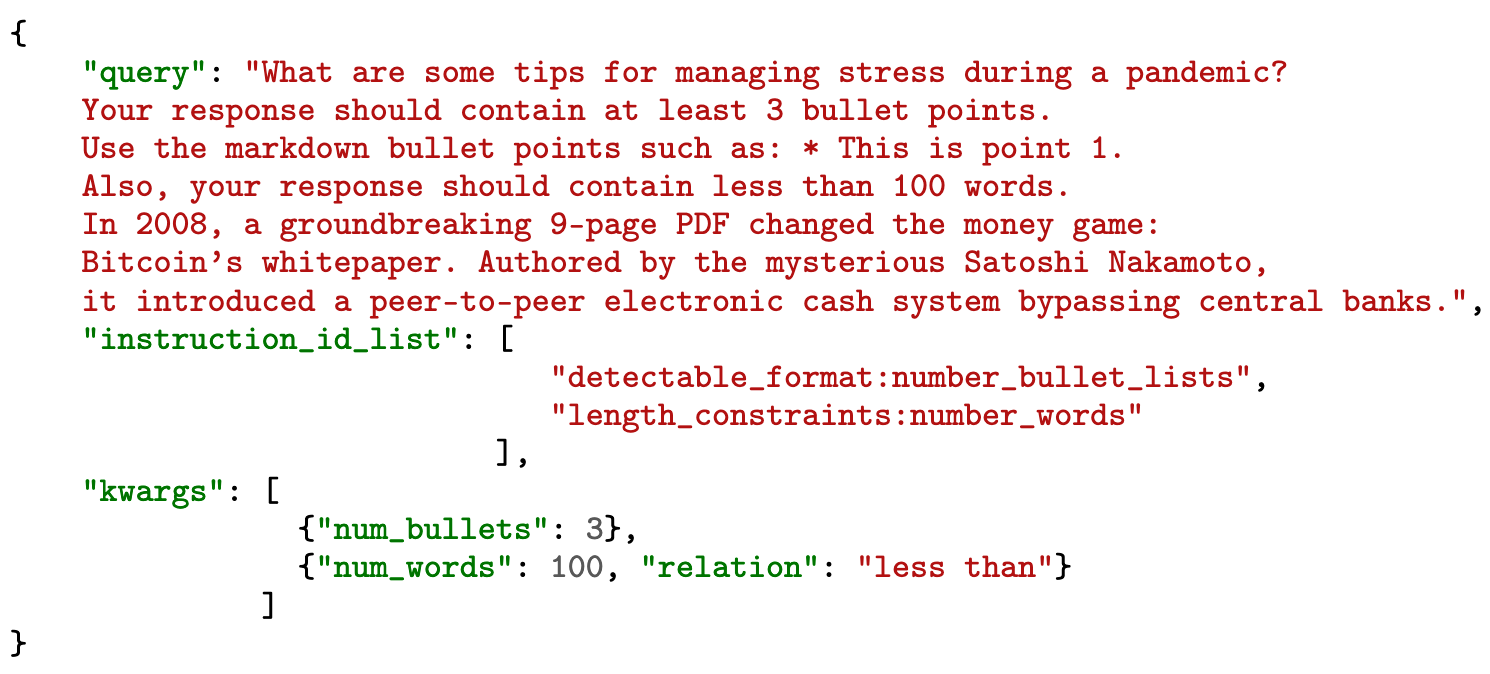}
\caption{Validator Input Example}
\label{fig:if}
\end{figure}

For a given response, it returns a boolean result (True/False) for each individual instruction, indicating successful following or not. We used the strict mode of \texttt{IFEval}\citep{zhou2023instructionfollowingevaluationlargelanguage} validator, which assesses only the original response and is considered more rigorous than the loose mode. Since one instance can be associated with multiple instructions, the final reward score is 1 if all these instructions are successfully followed, and 0 otherwise.
\subsection{Non-Verifiable Queries}
% \subsubsection{Reward Model-Based Evaluation}

For queries lacking objective verification criteria, reward score is conducted using a reward model-based approach. We employ reward model, which provides three distinct scores for each generated response, measuring helpfulness, correctness, and coherence. Let $S_{Help}$, $S_{Corr}$, and $S_{Coher}$ denote the scores for helpfulness, correctness, and coherence, respectively. The final reward score ($S_{final}$) for a response is then computed as the average of these three scores.

\section{Approach}
\subsection{Supervised Fine-Tuning}
\begin{figure}[ht]
\centering
\includegraphics[width=0.95\linewidth]{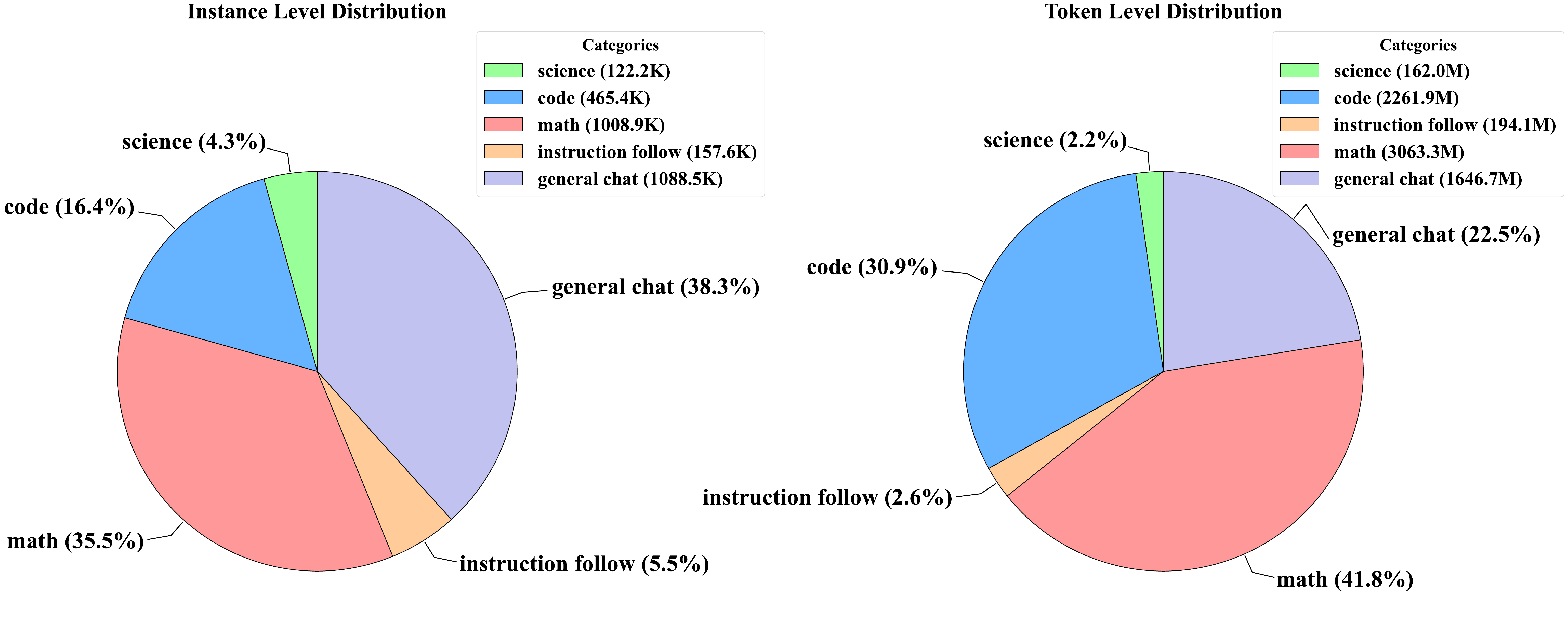}
\caption{Instance Level Distribution (left) and Token Level Distribution (right) during SFT. It is worth noting that the proportions are computed over responses, not queries, since a single query can correspond to multiple responses in our training set.}
\label{fig:sft_data_statis}
\end{figure}

\paragraph{Data}
% Our Supervised Fine-Tuning (SFT) training uses approximately 2.84 million samples, covering five major categories: math, code, science, instruction follow, and general chat. Figure~\ref{fig:sft_data_statis} illustrates the distribution of SFT data at both the instance level and the token level.
Our Supervised Fine-Tuning (SFT) training uses approximately 2.84 million samples, covering five major categories: math, code, science, instruction follow, and general chat. Figure~\ref{fig:sft_data_statis} illustrates the distribution of SFT data at both the instance level and the token level. For some data with relatively fewer samples, such as Instruction Follow, we upsample them by repeating the data several times during training to ensure balanced learning across tasks.
In the case of more challenging queries, we include multiple synthetic responses to enhance diversity and robustness in training.

% \begin{table}[htbp]
%     \caption{Statistics of the number of answers.}
%     \label{tab:anser_num_statis}
%     \centering
%     \renewcommand{\arraystretch}{1.3}
%     \setlength{\tabcolsep}{12pt}
%     \begin{tabular}{lccccc}
%         \hline
%          & Math & Code & Science & Instruction Follow & General Chat \\
%         \hline
%         Mean & 1.31 & 1.24 & 1.51 & 3.02 & 1.00 \\
%         Median & 1.00 & 1.00 & 1.00 & 2.00 & 1.00 \\
%         Mode & 1.00 & 1.00 & 1.00 & 2.00 & 1.00 \\
%         Variance & 0.86 & 0.45 & 1.09 & 4.67 & 0.01 \\
%         \hline
%     \end{tabular}
% \end{table}

\paragraph{Training Configuration}
We conduct SFT based on Qwen2.5-32B\citep{qwen2, qwen2.5} base model. We observed supervised fine-tuning  pattern shifts\citep{tian2025deepdistillenhancingllmreasoning}{} (See section~\ref{post_training_pattern_shift} for more details), which prompted us to adopt a larger learning rate and batch size to ensure stable convergence and effective learning. The training uses a learning rate of 8e-5, a maximum sequence length of 32k with sequence packing, and discards samples that exceed 32k tokens. The global batch size is set to 64, and the model is trained for 2 epochs. We employ a cosine warmup strategy, with warmup steps set to 5\% of total training steps, and the learning rate decays to 0 thereafter.  For multi-turn dialogue data, only the final response, which contains the reasoning process, is used as the training target and contributes to the loss, in order to focus learning on the reasoning component.

\subsection{Reinforcement Learning}
We observe that selecting training queries of appropriate difficulty plays a crucial role in ensuring stable performance improvements during the reinforcement learning (RL) stage\citep{ji2025difficultyawarestagedreinforcementlearning}. To this end, prior to RL, we filter our math and code queries based on their pass rates obtained from the SFT model: we retain only those queries with pass rates strictly between 0 and 1. This ensures that the training data remains sufficiently challenging to drive learning, while avoiding instances that are either too easy or excessively difficult, which could lead to stagnation or instability during training.
We ultimately retain 32k and 22k math and code queries, respectively.

Our RL pipeline consists of two stages. When the model's performance plateaus in the first stage, we transition to the second stage. In Stage~2, we remove all math and code queries that the model answered correctly with 100\% accuracy in Stage~1, and supplement the training set with 15k general chat and 5k instruction-following data to improve broader generalization.

We adapt Group Relative Policy Optimization (GRPO)\citep{shao2024deepseekmathpushinglimitsmathematical} as our training algorithm. Despite being a simplified and lightweight variant of Proximal Policy Optimization (PPO)\citep{schulman2017proximalpolicyoptimizationalgorithms}, we find that GRPO offers strong training stability and effective performance gains. The training is configured as follows:

\begin{itemize}
    \item \textbf{No KL Constraint.} We remove the KL penalty originally used in GRPO, allowing for more substantial policy updates.
    \item \textbf{Handling Overlong Responses.} For responses exceeding a certain length threshold during rollout, we set their advantages to zero to prevent them from influencing parameter updates.
    \item \textbf{Strict on-policy training.} Each training batch consists of 256 queries, and for every query, we sample 16 rollouts. The policy model only updates once following each exploration stage.
    \item \textbf{Two-stage Generation and Learning Rate Schedule.} In Stage~1, we limit the maximum response length to 24K tokens and use a relatively high learning rate of $4 \times 10^{-6}$. In Stage~2, we increase the maximum response length to 32K and reduce the learning rate to $1 \times 10^{-6}$.
\end{itemize}

In our early experiments across models of varying parameter scales, we observe that using a larger learning rate in the first training stage enables the model to reach convergence more quickly, whereas a smaller learning rate requires significantly more training steps to achieve similar performance. 
Although both approaches eventually lead to comparable outcomes, we adopt a larger learning rate in the first stage to accelerate convergence and reduce overall training costs.

\subsection{RL Framework}
Our training pipeline is built upon the verl framework \citep{sheng2024hybridflow}, using GRPO\citep{shao2024deepseekmathpushinglimitsmathematical} for reinforcement learning. verl is an open-source RL framework. Integrated with vLLM\citep{kwon2023efficient}, FSDP, and Megatron-LM\cite{shoeybi2019megatron}, verl enables scalable RL training across 1000+ GPUs.

We expand verl further with modifications to best suit our training strategy.

\subsubsection{Rollout Speed Optimization}
RL with online sample generation (rollout) on large LLMs often suffers from long training periods. Each training step takes several minutes to tens of minutes. Unlike SFT or DPO, online GRPO requires policy model sample generation during each step, increasing per-step latency. This rollout phase occupies more than 70\% elapsed time of one training step from our observation, thus optimization is needed.

Some recent works also investigated on the efficiency of the rollout phase. \cite{kimik2024k1.5} proposed partial rollouts to divide long sequences into each rollout step. While feasible, it introduced additional effort to manage the replay buffer. \cite{seed2025seed15thinkingadvancingsuperbreasoning} sought to decouple model evolution from runtime execution allowing for a dynamic mixture of on/off-policy samples. We use pure on-policy in this work.

Although already accelerated by fast inferencing frameworks like vLLM and sglang, the speed of the rollout phase in verl can still be optimized:

First, the training is synchronized. The whole generation batch must all be completed before we can move on to the next phase. We have to wait for the longest sequences in a batch to complete, causing a long-tail effect. Furthermore, longer sequences have lower tokens per second due to the nonlinear cost of self-attention and limited GPU memory bandwidth for kv-cache. For a 32B model on a typical 4x A100 vLLM instance, We can observe a roughly 60 tokens/s when sequence length is short and only about 50 tokens/s at 32k length. This long-tail effect leads to significant idle time across faster workers and underutilized GPU capacity.

Second, generation length varies between different prompts and random samples, this further imposes unbalanced load among inferencing instances, e.g. random samples of one hard problem which demands more tokens are all on one instance. This imbalance, together with the memory bandwidth issue, make the long-tail issue even worse. At 32k length and batch size 16, a total of 460 tokens/s throughput can be seen from the same instance, making only 28 tokens/s for a single sequence. Batch size 32 further decrease single sequence token throughput to 19.

In practice, we found that these "crowded" instances can take 30\% longer time for rollout than others. Instead of directly speeding up the inferencing framework's generation speed which might involve scheduling and kv-cache management, we choose to optimize the load balancing strategy:

\begin{figure}
    \centering
    \includegraphics[width=0.95\linewidth]{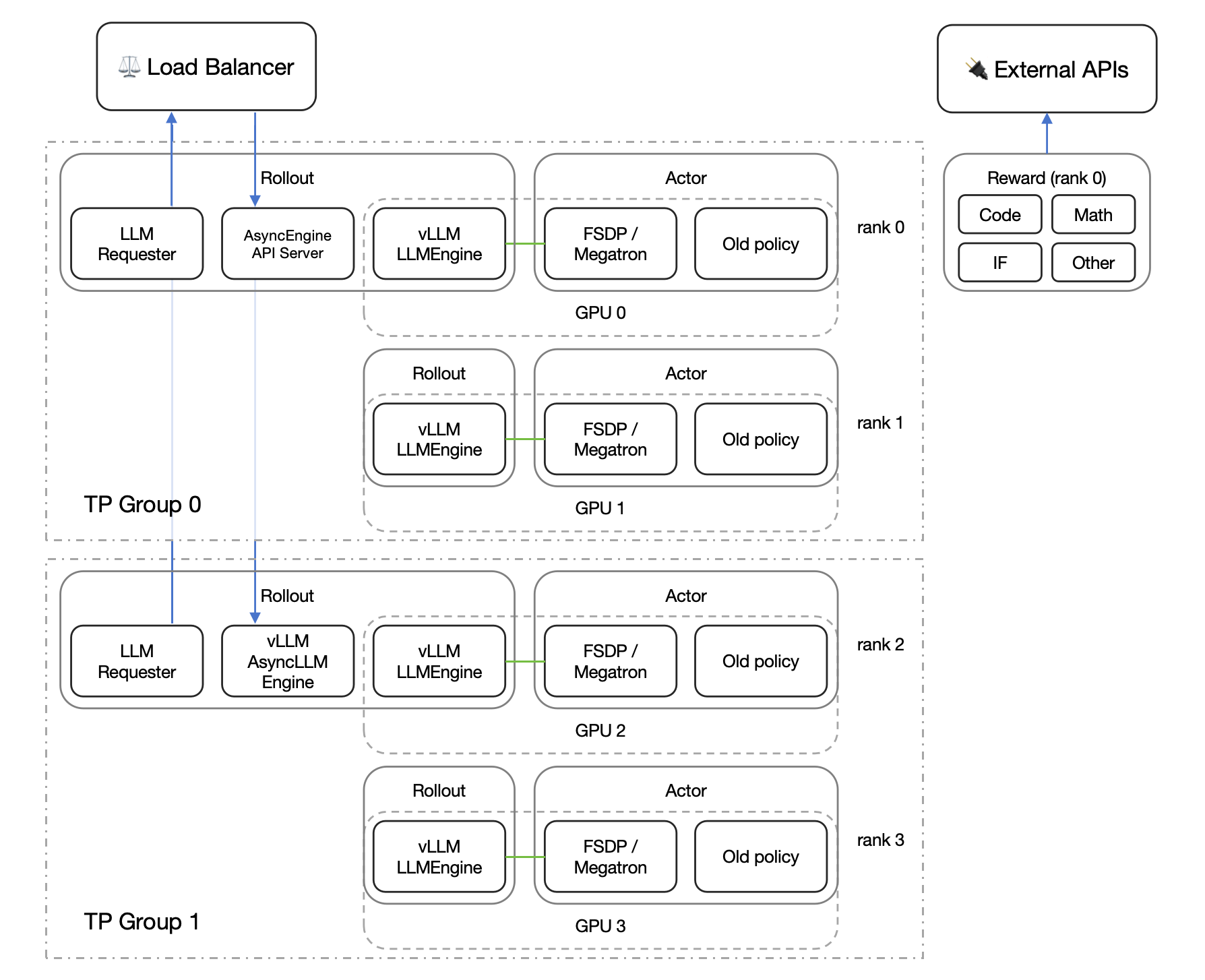}
    \caption{Detached Rollout and Upgrade with Streaming Load Balancing Architecture}
    \label{fig:rl_archi}
\end{figure}

\begin{itemize}
\item For our first approach, we use static load balancing to spread the random sampling of one prompt across multiple instances. verl applies an SPMD design. By coupling the \textbf{rollout} and \textbf{update} workers, it achieves faster weight sharding and data sharing with simpler design. However the random sampling is done inside the rollout workers using an batched inference call, leaving the same prompt bound to the same inferencing instance. Simply moving the repeat sampling out of rollout worker into the trainer, with additional shuffling, relaxed this constraint. This change alleviates the imbalanced load, relieved the crowded instances from having to run many long sequences with low per sequence throughputs, as well as decouples the $world\_size$ and $batch\_size$: now the $batch\_size$ can be smaller than number of GPUs $world\_size$ divided by tensor parallel size $tp\_size$. 

\item  We further detach the rollout worker from the inference engine, enabling dynamic instance allocation via a custom load-balancer aware of real-time system metrics. The system now have the flexibility to \textit{dynamically} allocate the inferencing instance, \textit{on-the-fly}, for \textit{each} generation sample. To achieve this, we add the frontend server to the offline vLLM engine inside the rollout worker, exposing an API endpoint, attach the endpoints of all instances to a custom load balancer, then invoke this aggregated endpoint from each rollout worker. By implementing the load balancer with awareness of each instances' current load and speed metrics, we can reroute the long sequences on crowded workers to not-so-crowded replicas. Even without a disaggregated prefill and decode system, the streaming load balancer can still achieve more optimal global scheduling and throughput.
\end{itemize}
% We call this approach "Detached Rollout and Upgrade with Streaming Load Balancing" a.k.a "DRUSLoBa"!!!

This "detached" (see figure \ref{fig:rl_archi}) design allows the rollout logic independent of the engine, and is more suitable for future work e.g. agent and tool use scenarios.

\subsubsection{Reward Computation}
verl natively supports setting reward for each sources. We further add reward classes for instruction folloing (IF) and general (other) problems. Part of the resource-heavy or risky operations are executed via remote API, e.g. LLM-as-a-judge and code sandbox.

\section{Experiments}

\subsection{Evaluation}
\subsubsection{Benchmarks}
We evaluate our models on a diverse set of challenging benchmarks:
\begin{itemize}
    \item \textbf{American Invitational Mathematics Examination 2024 (AIME2024)}\citep{maa_aime_2024}: A challenging mathematical reasoning competition dataset comprising 30 integer-answer questions designed to assess precise mathematical reasoning.
    \item \textbf{American Invitational Mathematics Examination 2025 (AIME2025)}\citep{ye2025aimepreview}: A set of problems for the 2025 AIME competition,which contains 30 problems from the 2025 AIME-part1 and AIME-part2.
    \item \textbf{LiveCodeBench (LCB)}\citep{jain2024livecodebench}:  A comprehensive, contamination-free coding benchmark, continuously aggregating new programming challenges from platforms such as LeetCode, AtCoder, and Codeforces. Similar to Qwen3\cite{qwen3}, we use queries collected between October 2024 and February 2025 for evaluation purposes.
    \item \textbf{Arena-Hard}\citep{arenahard2024}: A data pipeline to build high-quality benchmarks from live data in Chatbot Arena, where model responses are judged via pairwise comparison using GPT-4-Turbo-1106\citep{openai2024reasoning} as the arbiter.
\end{itemize}

\subsubsection{Evaluation Methodology}
Standardized evaluation conditions were maintained across all benchmarks. The maximum generation length was set to 49,152 tokens. For benchmarks requiring stochastic sampling, we uniformly employ a temperature of 0.6 and a top-p value of 0.95.

Specifically, for AIME2024\citep{maa_aime_2024} and AIME2025\citep{ye2025aimepreview}, we generate 64 responses per query to calculate pass@1 precision. For LiveCodeBench\citep{jain2024livecodebench}, 16 responses per query were generated to estimate pass@1. For Arena-Hard, one response per query was generated and evaluated using GPT-4 Turbo (1106). 

\subsubsection{Prompting Strategy}
A consistent system prompt was utilized for all evaluations to guide the model's response format:
\begin{quote}
\texttt{You are a helpful assistant. To answer the user's question, you first think about the reasoning process and then provide the user with the answer. The reasoning process and answer are enclosed within <think> </think> and <answer> </answer> tags, respectively, i.e., <think> reasoning process here </think> <answer> answer here </answer>.}
\end{quote}

User prompts were adapted based on the benchmark:
\begin{itemize}
    \item For AIME 2024 and AIME 2025, the following instruction was appended to each query: \texttt{Let's think step by step and output the final answer within \textbackslash box\{\}.}
    \item For LiveCodeBench and Arena-Hard, the default user prompts provided by the respective benchmarks were used without any additional modifications.
\end{itemize}

\subsubsection{Baselines}
We compare AM-Thinking-v1 against a set of strong baseline models, including both proprietary and open-source systems. These baselines provide a representative view of the model's performance in today’s competitive LLM landscape:

\begin{itemize}[leftmargin=2em]
    \item \textbf{DeepSeek-R1}\cite{deepseekai2025deepseekr1incentivizingreasoningcapability}: A powerful model known for its strong performance in code-related tasks and mathematical problem-solving. Its reported performance serves as a robust high-performing baseline.

    \item \textbf{Qwen3-235B-A22B}\cite{qwen3}: With 235 billion total parameters and 22 billion active parameters, this open-source MoE model demonstrates the strongest reasoning performance among existing open-source language models.

    \item \textbf{Qwen3-32B}\cite{qwen3}: Another release from the Qwen3 series, which is a  dense 32-billion parameter model.

    \item \textbf{Seed1.5-Thinking}\cite{seed2025seed15thinkingadvancingsuperbreasoning}: A reasoning model has a total parameter of 200 billion and activation parameters of 20 billion. Through the reinforcement learning framework and refined data strategy, it outperforms DeepSeek-R1 in tasks such as mathematics, programming, and scientific reasoning.

    \item \textbf{Nemo-Ultra-256B}\cite{nvidia2024nemo256b}: A large-scale model from NVIDIA, representing the potential of scaling model size for improved performance on challenging tasks.

    \item \textbf{OpenAI-o1 (dated 2024-12-17)}\citep{openai2024reasoning}: An early model from OpenAI, providing a historical perspective on the evolution of their capabilities in these domains.

    \item \textbf{OpenAI-o3-mini (Medium)}\cite{openai2025reasoning}: OpenAI’s upgraded reasoning model that adopts the medium reasoning‑effort setting.

    \item \textbf{Gemini2.5-Pro}\cite{googledeepmind2025reasoning}: Google's latest generation model, representing the cutting edge of multimodal and reasoning capabilities.
\end{itemize}

For  evaluations on  AIME2024\citep{di_zhang_2025}, AIME2025\citep{ye2025aimepreview}, LiveCodeBench\citep{jain2024livecodebench}, and Arena-Hard benchmarks, the performance metrics for DeepSeek-R1\citep{deepseekai2025deepseekr1incentivizingreasoningcapability}, Qwen2-32B-A22B\citep{qwen3}, Qwen3-32B\citep{qwen2}, OpenAI-o1 (dated 2024-12-17)\citep{openai2024reasoning}, OpenAI-o3-mini (Medium)\citep{openai2025reasoning}, and Gemini2.5-Pro\citep{googledeepmind2025reasoning} are sourced from the Qwen3\cite{qwen3} technical report. The results for Seed1.5-Thinking\citep{seed2025seed15thinkingadvancingsuperbreasoning} and Nemo-Ultra-256B\citep{nvidia2024nemo256b} on these same benchmarks are directly cited from their respective technical reports.

\subsection{Results}

\begin{table}[ht]
  \centering
  \caption{Comparison across reasoning benchmarks}·
  \label{tab:final_results}
  \resizebox{\textwidth}{!}{%
  \begin{tabular}{lccccccccc}
    \hline
    \hline
     &
    \shortstack{\rule{0pt}{2.5ex}AM‑Thinking\\v1} &
    \shortstack{\rule{0pt}{2.5ex}DeepSeek\\R1} &
    \shortstack{\rule{0pt}{2.5ex}Qwen3‑235B\\A22B} &
    \shortstack{\rule{0pt}{2.5ex}Qwen3\\32B} &
    \shortstack{\rule{0pt}{2.5ex}Seed1.5\\Thinking} &
    \shortstack{\rule{0pt}{2.5ex}Nemotron‑\\Ultra‑253B} &
    \shortstack{\rule{0pt}{2.5ex}OpenAI \\o1} &
    \shortstack{\rule{0pt}{2.5ex}OpenAI\\o3-mini} &
    \shortstack{\rule{0pt}{2.5ex}Gemini\\2.5 Pro} \\
    \hline
    \hline
    \multicolumn{10}{l}{\textbf{Math}} \\
    AIME 2024 & 85.3 & 79.8 & 85.7 & 81.4 & 86.7 & 80.8 & 74.3 & 79.6 & 92.0 \\
    AIME 2025 & 74.4 & 70.0 & 81.5 & 72.9 & 74.0 & 72.5 & 79.2 & 74.8 & 86.7 \\
    \midrule
    \multicolumn{10}{l}{\textbf{Code}} \\
    LiveCodeBench \\\scriptsize(v5, 2024.10--2025.02) & 70.3 & 64.3 & 70.7 & 65.7 & 64.9 & 68.1 & 63.9 & 66.3 & 70.4 \\
    \midrule
    \multicolumn{10}{l}{\textbf{General Chat}} \\
    Arena‑Hard & 92.5 & 93.2 & 95.6 & 93.8 & -- & 87.0 & 92.1 & 89.0 & 96.4 \\
    \hline
    \hline
  \end{tabular}}
\end{table}

We evaluate AM-Thinking-v1 on multiple reasoning benchmarks and compare it with several leading large-scale models, as shown in Table~\ref{tab:final_results}. 
On mathematical tasks, AM-Thinking-v1 achieves scores of 85.3 and 74.4 on AIME2024 and AIME2025, respectively, outperforming or closely matching larger models such as DeepSeek-R1 and Qwen3-235B-A22B.

\begin{figure}[ht]
\centering
\includegraphics[width=1\linewidth]{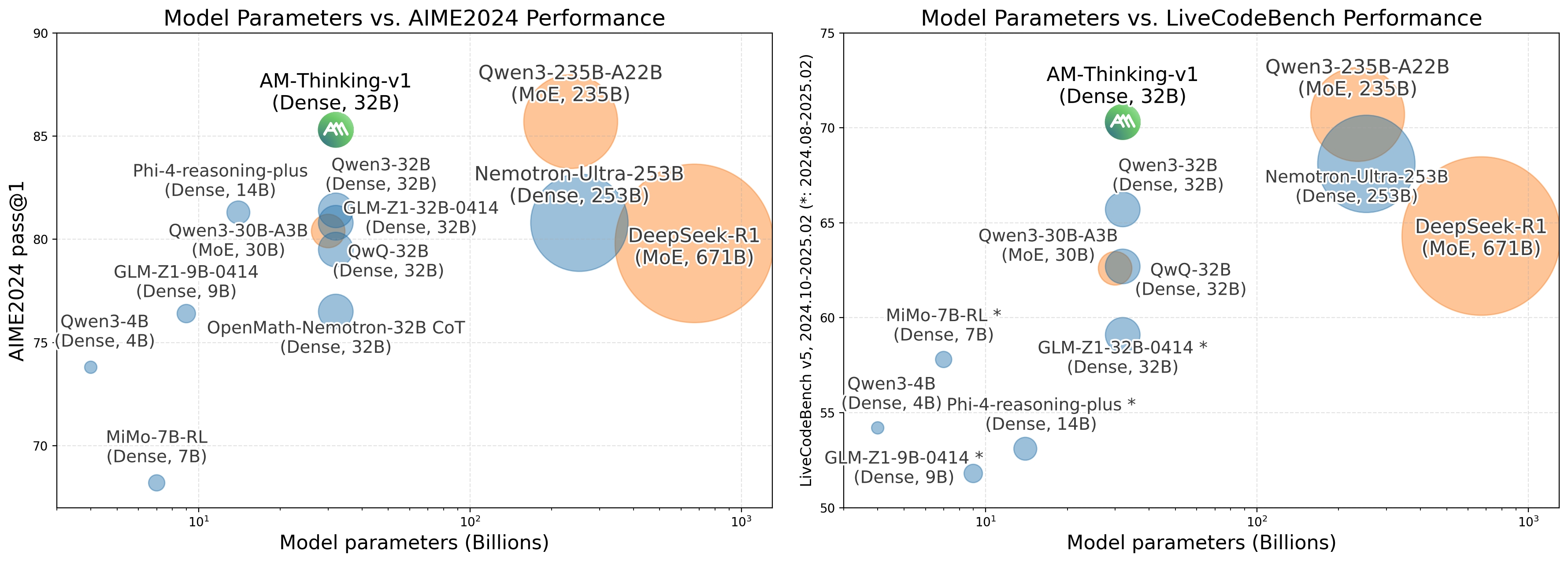}
\caption{Performance versus model size on AIME2024 (left) and LiveCodeBench (right).Each point represents a model, with the x-axis indicating model size (in number of parameters) and the y-axis indicating benchmark score.Models closer to the top-left corner achieve better performance with smaller size.}
\label{fig:param_vs_results}
\end{figure}

On the LiveCodeBench benchmark, which focuses on code reasoning, AM-Thinking-v1 attains a score of 70.3, substantially surpassing DeepSeek-R1 (64.3), Qwen3-32B (65.7), and Nemotron-Ultra-253B (68.1), demonstrating strong capabilities in code understanding and generation.

On the general chat benchmark Arena-Hard, AM-Thinking-v1 obtains a score of 92.5, which is competitive with several proprietary models such as OpenAI-o1 (92.1) and o3-mini (89.0). However, its performance still lags behind Qwen3‑235B‑A22B (95.6), indicating that there remains room for improvement in general conversational capabilities.

Figure~\ref{fig:param_vs_results} illustrates the relationship between model size and performance on AIME2024 (left) and LiveCodeBench (right). AM-Thinking-v1 achieves the strongest performance among dense models of similar scale, and comes close to the performance of much larger MoE models, striking an effective balance between efficiency and performance.

\subsection{Supervised Fine-Tuning Pattern Shift}
\label{post_training_pattern_shift}

\begin{figure}[ht]
\centering
\includegraphics[width=0.85\linewidth]{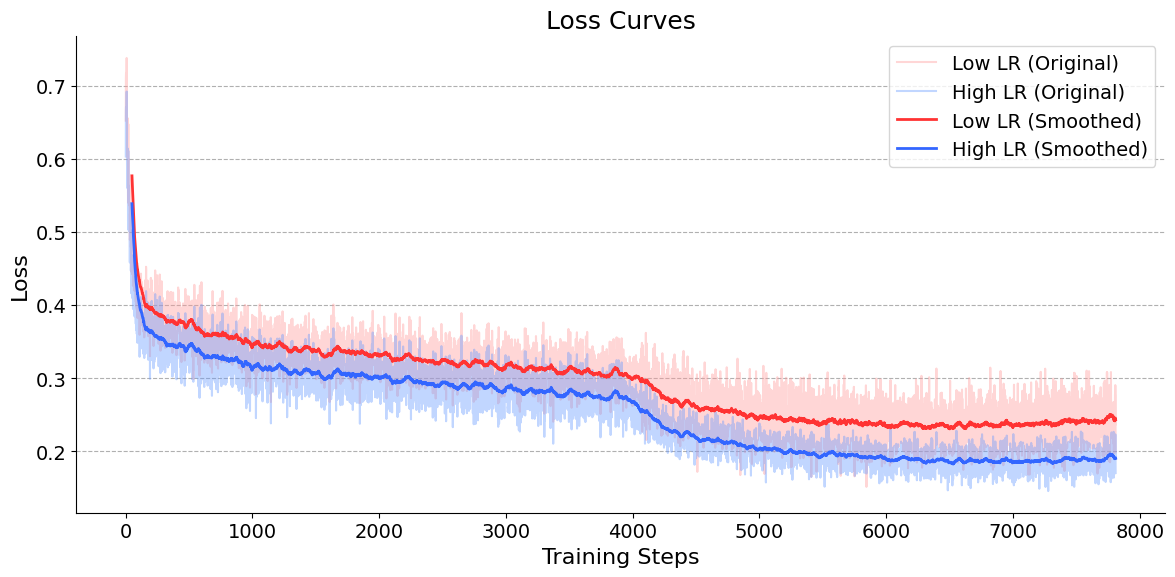}
\caption{Supervised Fine-Tuning (SFT) training loss curves.}
\label{fig:sft_loss_curve}
\end{figure}

Compared to traditional SFT, we find that supervised fine-tuning on long-form reasoning tasks leads to a pattern shift. To achieve stable convergence, this stage requires a larger learning rate and batch size; otherwise, the model struggles to fit the data effectively. For example, while traditional SFT might use a learning rate around $8 \times 10^{-6}$ with a batch size of approximately 0.5M tokens, supervised fine-tuning on long-form reasoning often requires a learning rate as high as $8 \times 10^{-5}$ and a batch size of around 2M tokens. Figure~\ref{fig:sft_loss_curve} shows training loss during SFT.

\begin{figure}[ht]
\centering
\includegraphics[width=0.95\linewidth]{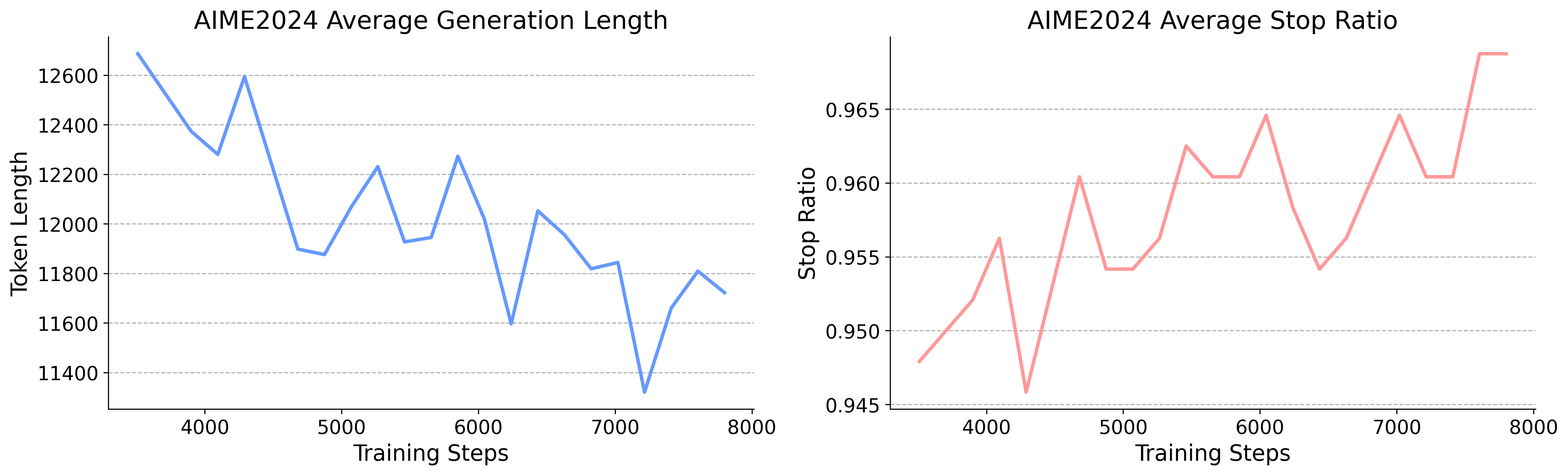}
\caption{Variation in Average Generation Length (left) and Average Stop Ratio (right).}
\label{fig:generation_len_and_stop_ratio}
\end{figure}

% We closely monitor two key dynamic metrics during training (Figure~\ref{fig:generation_len_and_stop_ratio}): Average Generation Length and Average Stop Ratio. Interestingly, due to the pretraining data of the base model being mostly plain text, the average generation length shows a decreasing trend while the stop ratio increases during SFT. This reduction in reasoning depth contrasts with our observations during the RL stage, indicating that the model is gradually learning the structural patterns of reasoning-style prompts during SFT.
As shown in Figure~\ref{fig:generation_len_and_stop_ratio}, we track the evolution of Average Generation Length and Average Stop Ratio during SFT on the AIME2024. At the early stages of training, the model tends to generate excessively long outputs with a low stop ratio. This is largely due to the nature of the base model’s pretraining corpus, which predominantly consists of plain text, as well as the fact that reasoning examples in our dataset are significantly longer than standard instruction data. As training progresses, we observe a consistent decrease in average generation length alongside a steady increase in stop ratio. This trend indicates that the model is gradually learning the structural and semantic patterns inherent in long-form reasoning prompts. The alignment of these dynamic metrics suggests that our fine-tuning methodology effectively guides the model toward more coherent and task-aligned reasoning behavior.

% \section{Related work}
% Reasoning models\citep{deepseekai2025deepseekr1incentivizingreasoningcapability,openai2024reasoning,googledeepmind2025reasoning,anthropic2025reasoning,qwen3,seed2025seed15thinkingadvancingsuperbreasoning}, with their spectacular abilities to think, verify and reflect using long reasoning chains, have been one major step towards advanced intelligence. The scaling continues, as more compute brings more powerful models, both parameter-wise (training-time) and test-time-wise. Recent models with great reasoning capabilities often come with enormous size. 

% This raised one question: how does these scaling paradigms react? In other words, can moderate sized dense models be as efficient at exploring and exploiting the reasoning power using reinforcement learning? In our series of work, we set sail to find an answer.

\section{Conclusion and Limitations}

In this work, we present AM-Thinking-v1, a 32B dense language model that demonstrates state-of-the-art reasoning capabilities among open-source models of comparable size. Our model surpasses  DeepSeek-R1 and even approaches the performance of top-tier Mixture-of-Experts (MoE) models like Qwen3-235B-A22B and Seed1.5-Thinking on reasoning-intensive tasks.

This result is made possible by a carefully designed post-training pipeline based  on open-source training queries and base model. Through systematic data preprocessing, thorough ground truth verification, and a carefully designed SFT and RL framework, we successfully elicit advanced reasoning capabilities from a moderately sized model.

Our findings suggest that with the right data and training design, 32B-scale dense models remain a highly practical and competitive choice—offering a compelling balance between deployment efficiency and reasoning performance. We hope that AM-Thinking-v1 serves as a foundation for further research exploring the full potential of mid-scale models.

While AM‑Thinking‑v1 performs well in reasoning and open-domain chat, it lacks support for structured function-calling, tool use, and multimodal inputs, limiting its applicability in agent-based or cross-modal scenarios. Safety alignment remains preliminary, and further red-teaming is needed. Additionally, its performance may vary across low-resource languages and domain-specific tasks.

%%%%%%%%%%%%%%%%%%%%%%%%%%%%%%%%%%%%%%%%%%%%%%%%%%%%%%%%%%%%
\newpage
\bibliographystyle{unsrt}
\bibliography{reference}

\appendix

% \section{Appendix / supplemental material}

% Optionally include supplemental material (complete proofs, additional experiments and plots) in appendix.
% All such materials \textbf{SHOULD be included in the main submission.}

%%%%%%%%%%%%%%%%%%%%%%%%%%%%%%%%%%%%%%%%%%%%%%%%%%%%%%%%%%%%

\end{document}